\documentclass[twocolumn, 10pt]{article}

\usepackage[utf8]{inputenc}
\usepackage[T1]{fontenc}
\usepackage[margin=0.75in]{geometry}
\usepackage{amsmath, amssymb, amsthm}
\usepackage{graphicx}
\usepackage{booktabs}
\usepackage{xcolor}
\usepackage{lipsum}
\usepackage{hyperref}
\usepackage{tikz}
\usepackage[blocks]{authblk}
\usetikzlibrary{arrows.meta, positioning, shadows}
\usepackage[numbers]{natbib}
\definecolor{darkblue}{rgb}{0.0, 0.0, 0.5}
\hypersetup{colorlinks=true, linkcolor=darkblue, citecolor=darkblue, urlcolor=darkblue}

\title{\textbf{From Rubble Simulation to Active Magnetic Mapping: Quantum Sensing for Disaster Response}}
\author[1]{Samuel Tovey\thanks{\texttt{samuel.tovey@quantum-brilliance.com}}}
\author[1]{Stefan Prestel\thanks{\texttt{stefan.prestel@quantum-brilliance.com}}}
\author[2]{Hiroshi Yamauchi\thanks{\texttt{hiroshi.yamauchi@g.softbank.co.jp}}}

\affil[1]{Quantum Brilliance}
\affil[2]{SoftBank Corp.}

\begin{document}

\maketitle

\begin{abstract}
	Locating survivors of building collapses within the first 72 hours is a critical challenge in disaster response, and existing sensing modalities provide only partial information about the structure beneath the rubble.

This paper proposes drone-based quantum magnetometry as a complementary modality and develops a simulation pipeline spanning rubble physics, sensor-array deployment, and active spatial reconstruction.
We use Unreal Engine to generate a steel-reinforced concrete parking-garage collapse and compute the induced magnetic field via a per-triangle dipole approximation, establishing that meaningful magnetic structure is recoverable in the sub-pT to sub-nT range from roughly 1 m above the roofline.
Then, we feed sparse multi-sensor samples into a Gaussian Process Regression back-end driven by Bayesian active sampling and validate the pipeline across multiple independent collapse realizations; a three-sensor array optimizes the trade-off between gradient resolution and UAV payload constraints, and active sampling reaches peak structural correlation in roughly $100$ samples.
Together, these results indicate that quantum-grade sensing could become a useful tool for drone-based structural analysis and potentially void detection in collapsed buildings.

\end{abstract}

\section{Introduction}
\label{sec:introduction}
Locating and rescuing survivors of building collapse within the first 72 hours of an incident is a critical challenge faced by governments around the world.
Time-to-rescue analyses of post-earthquake structural entrapment~\citep{macintyre06a}, complemented by reviews of earthquake medical complications~\citep{bartels12a} and broader entrapment-survivability studies~\citep{ngwenyama25a}, have established that survival probability decreases sharply after the first two days, driven by dehydration, crush syndrome, and exposure.
Rescue operations consequently employ a stack of complementary sensing modalities: multimodal rubble-sensing systems combining audio and tactile cues~\citep{zhang18a}, search cameras and snake-arm endoscopes deployed by Urban Search and Rescue (USAR) teams~\citep{casper03a, murphy14a}, thermal imaging through gaps in the rubble, trace detection of human volatile organic compounds and exhaled CO$_{2}$~\citep{agapiou15a}, ground-penetrating radar for void localization~\citep{crocco14a}, and integrated sensor-fusion approaches to in-rubble void detection~\citep{hu19a}, all increasingly supported by aerial platforms providing situational awareness from above~\citep{erdelj17a}.
Each of these modalities, however, provides only partial information about the structure of the rubble itself.
The introduction of stable quantum sensors~\citep{wang24a, herb25a, stolz24a} has the potential to greatly expand the amount of information we can recover from a disaster site. 
One promising approach comes in the form of quantum magnetometry: the sensing of fluctuations in a magnetic field via a quantum mechanical process.
Given sufficient sensitivity, such sensors could detect and characterize the subtle fluctuations in magnetic fields arising from common building materials such as steel-reinforced concrete, narrowing search areas and informing the safe removal of debris.

An adjacent research direction concerns how magnetic measurements should be acquired and integrated by an autonomous platform.
Magnetic Simultaneous Localization and Mapping (MagSLAM)~\citep{vallivaara11a, kok18a, solin18a} treats local ferromagnetic anomalies as stable, passive landmarks to solve the related problem of infrastructure-free positioning in GPS-denied environments.
We instead focus on the complementary task of efficiently reconstructing the underlying field given a known platform pose, an axis that has received less attention but is central to converting raw drone overflights into actionable structural maps.
Magnetic-anomaly methods of either flavor have historically relied on classical magnetometers, fundamentally limiting achievable map resolution.
Recent advances in solid-state quantum sensing, e.g., Nitrogen-Vacancy (NV) centers in diamond, provide a high-performance alternative: low-drift, high-bandwidth vector magnetometry at room temperature with high sensitivity~\citep{sengottuvel22a}.
Even with these sensing advantages, the spatial sampling of magnetic fields by autonomous agents remains a bottleneck, since exhaustive grid searches are inefficient in both battery life and computational overhead, and will be limited to small sample areas.

This paper bridges the physical and algorithmic sides of this problem within a single integrated pipeline.
We first model a realistic building collapse and the associated induced magnetic field, establishing the sensitivity envelope required of an onboard magnetometer for structure to be discernible from a few meters above the rubble.
We then show that a Gaussian Process Regression back-end driven by Bayesian active sampling~\citep{frazier18a} reconstructs continuous magnetic topologies from sparse, multi-sensor samples with high fidelity, and that this conclusion holds across multiple independent collapse realizations.

\section{Theory}
\label{sec:theory}
We split the formalism into two pieces matching the two contributions of this work: a physical model that converts a rubble mesh into a magnetic field, and a Gaussian-process reconstruction that turns sparse field samples into a continuous structural map.

\subsection*{Dipole model of the rubble field}
Each triangle of the rubble mesh is treated as a thin slab of steel-reinforced concrete whose magnetic moment is induced by the local Earth field. 
Treating the slab as having uniform thickness $t$ absorbed into the constant prefactor we use for the moment, the effective moment vector $\mathbf{m}_i$ of the $i^{\text{th}}$ triangle is
\begin{equation}
    \mathbf{m}_i = A_i\,\chi\,\mathbf{B}_{\text{earth}},
\end{equation}
with $A_i$ the triangle area, $\chi = \mu_r - 1$ the dimensionless susceptibility, and $\mathbf{B}_{\text{earth}}$ the inducing Earth field; the implicit $t/\mu_0$ factor of a fully dimensional moment is held constant across triangles and absorbed into the sum below. 
The total field vector at an arbitrary point $\mathbf{p}$ is the dipole sum
\begin{equation}
    \mathbf{B}(\mathbf{p}) = \frac{\mu_0}{4\pi}\sum_{i=1}^{N}\frac{3\hat{\mathbf{r}}_i(\mathbf{m}_i \cdot \hat{\mathbf{r}}_i) - \mathbf{m}_i}{|\mathbf{r}_i|^3},
\end{equation}
where $\mathbf{r}_i = \mathbf{p} - \mathbf{c}_i$ runs from the center of triangle $i$ to the evaluation point and $\hat{\mathbf{r}}_i$ is the corresponding unit vector.
Finally, we assume a dipole approximation throughout, which is only valid at a distance from the mesh.
For near-field applications, such as ground-based robots, a surface-integral formulation~\citep{singh00a, okabe79a} would be more appropriate.

\subsection*{Gaussian-process map reconstruction}
We reconstruct a continuous map of $\|\mathbf{B}\|$ on a horizontal operating plane from sparse drone samples using Gaussian Process Regression (GPR)~\citep{rasmussen04a}. 
The covariance is a Radial Basis Function (RBF) kernel plus a white-noise term,
\begin{equation}
    k(\mathbf{x},\mathbf{x}') = \sigma_f^2 \exp\left(-\frac{\|\mathbf{x}-\mathbf{x}'\|^2}{2l^2}\right) + \sigma_n^2\,\delta(\mathbf{x},\mathbf{x}'),
\end{equation}
giving a predictive mean $\mu(\mathbf{x})$ and standard deviation $\sigma(\mathbf{x})$ at every point. 
The next drone waypoint is chosen by an Upper Confidence Bound acquisition $\mathrm{UCB}(\mathbf{x}) = \mu(\mathbf{x}) + \kappa\,\sigma(\mathbf{x})$~\citep{frazier18a} in the pure-exploration limit $\kappa \to \infty$, equivalent to sampling at the location of maximum predictive uncertainty.

\section{Methodology}
\label{sec:methodology}
The simulation pipeline runs in three stages: a destructible building model produces a rubble mesh, the mesh is converted into a ground-truth magnetic field via the dipole model of Section~\ref{sec:theory}, and an autonomous mapping back-end reconstructs the field from sparse sensor readings along the drone's flight path. All computations were carried out on an NVIDIA H100 GPU through the JAX library~\citep{jax}.

\subsection*{Collapse simulation}
\begin{figure}
    \centering
    \includegraphics[width=\linewidth]{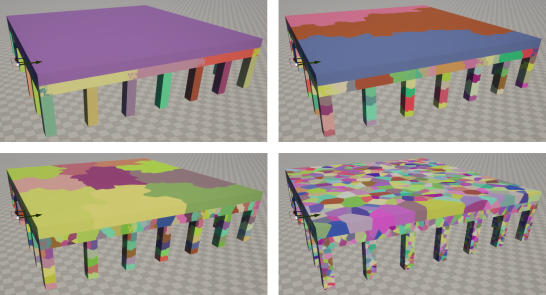}
    \caption{
        Fracture groups in the Unreal Engine Chaos simulation. 
        Larger groups split into smaller ones as the applied force grows.
    }
    \label{fig:groups}
\end{figure}

The building models were constructed in Unreal Engine~\citep{unrealengine} as a 28 m $\times$ 28 m $\times$ 5 m parking garage held up by $31$ columns, with a $0.5$ m thick concrete roof. 
All structural components are treated as a single steel-reinforced concrete material. 
Collapse is handled by the Chaos engine, which fractures the mesh via Voronoi tessellation under a sudden lateral impulse mimicking an earthquake (Figure~\ref{fig:groups}). 
Rather than instrument an explicit rebar lattice, we model the binding action of steel reinforcement by grouping Voronoi fractures into nested blocks, so that a slab first breaks into a few large fragments and only later disaggregates into smaller pieces. 
As the breakage parameters are seeded by a random variable, different simulations give rise to a range of plausible collapse geometries.

\subsection*{Ground-truth field parameters}
We evaluate the dipole model of Section~\ref{sec:theory} at parameters appropriate for a Tokyo deployment. 
The Earth field is taken from the NOAA geomagnetic model as $\mathbf{B}_{\text{earth}} = [30.0812, -4.1894, 35.6604] \cdot 10^{-6}$ T, in coordinates aligned to North-South, East-West, and Up-Down respectively~\citep{chulliat25a}. 
We use $\mu_r = 30$ as a conservative estimate for the bulk composite reinforced-concrete response derived fro the work of~\citet{he19a}.
For the sensitivity characterization we evaluate the field on a $250^3$ point grid spanning $40 \times 40 \times 10$ m; for the autonomous-mapping experiments we use a single horizontal plane at the drone's operating altitude.
The dipole approximation neglects remnant magnetization, mesh-to-mesh field coupling, and the survival of intact rebar after the concrete crumbles.
The first two effects add a noise-floor correction, and the third will likely result in a more structured field than is modeled here.

\subsection*{Drone-based sensing and reconstruction}
A rigid multi-sensor array carried by the drone samples the local field at every waypoint, and the resulting sparse readings are fitted with the GPR back-end of Section~\ref{sec:theory}. We evaluate array sizes $N \in \{1, 2, 3, 4\}$ with co-located sensors at a fixed $0.3$ m offset from the drone center. Three flight strategies are compared:
\begin{itemize}
    \item \textbf{Lawnmower}, a deterministic zig-zag path at a fixed stride.
    \item \textbf{Sine wave}, a continuous sinusoidal sweep across the search area.
    \item \textbf{Active sampling}, a pure-variance Bayesian acquisition seeded by $5$ random waypoints.
\end{itemize}
Reconstruction quality is measured by the Pearson correlation $\rho$ against the dipole ground truth. 
The Bayesian-active configuration is repeated for $20$ random seed-point realizations and reported as the mean and $95\%$ bootstrap confidence interval; the deterministic strategies are bit-reproducible on identical inputs.
The GPR reconstruction is performed in scikit-learn with $5$ random restarts of the hyperparameter optimization.
These restarts produce an error on the order of $10^{-8}$ and are therefore not a significant source of variability in the results.

\section{Results}
\label{sec:results}
We organise the results around the two contributions of this paper. 
We first present the magnetic field generated by our collapse simulation and the standoff range over which structure remains visible above the rubble. 
We then evaluate the autonomous mapping pipeline that recovers that structure from sparse drone samples.

\subsection*{Magnetic field of the collapsed building}
\begin{figure*}[t!]
    \centering
    \includegraphics[width=\linewidth]{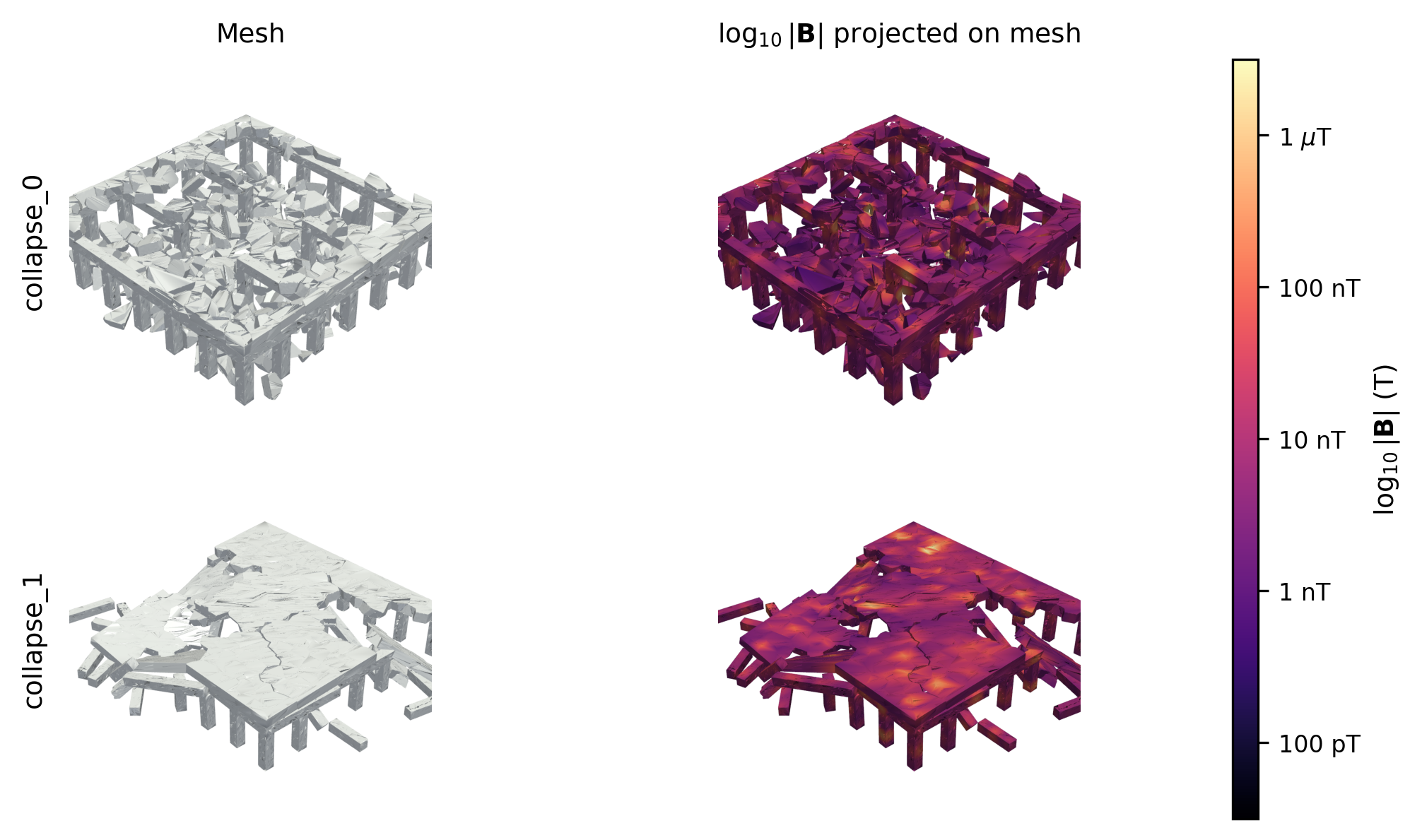}
    \caption{
        Two independent collapse realizations (top and bottom rows). 
        Left column: bare rubble mesh. Right column: dipole magnetic field projected on the mesh surface, $\log_{10}|\mathbf{B}|$ in T, with the color scale spanning 30 pT to 3 $\mu$T.
    }
    \label{fig:full-map}
\end{figure*}

The Unreal Engine simulation produces rubble meshes of the kind shown in the left column of Figure~\ref{fig:full-map}. 
Most of the load-bearing pillars remain standing while the roof fragments into grouped blocks: a visual signature of the fracture-group method described in Section~\ref{sec:methodology}, in which intact rebar prevents complete disaggregation of the slab. 
Evaluating the dipole model on each mesh and projecting the result onto the rubble surface gives the field overlays in the right column. 
Pillars and large roof panels with high triangle counts carry the strongest fields; loose, detached fragments are weakest.
Two field regimes are relevant to the deployment. The surface-projected fields shown here span approximately 30 pT to 3 $\mu$T near the rubble source region, characterizing the intensity at the dipole sources themselves; this is not what an aerial sensor would observe. The aerial stand-off slices relevant to UAV operation, presented below, instead exhibit structurally informative contrast in the sub-pT to sub-nT range.

\begin{figure*}[h!]
    \centering
    \includegraphics[width=\linewidth]{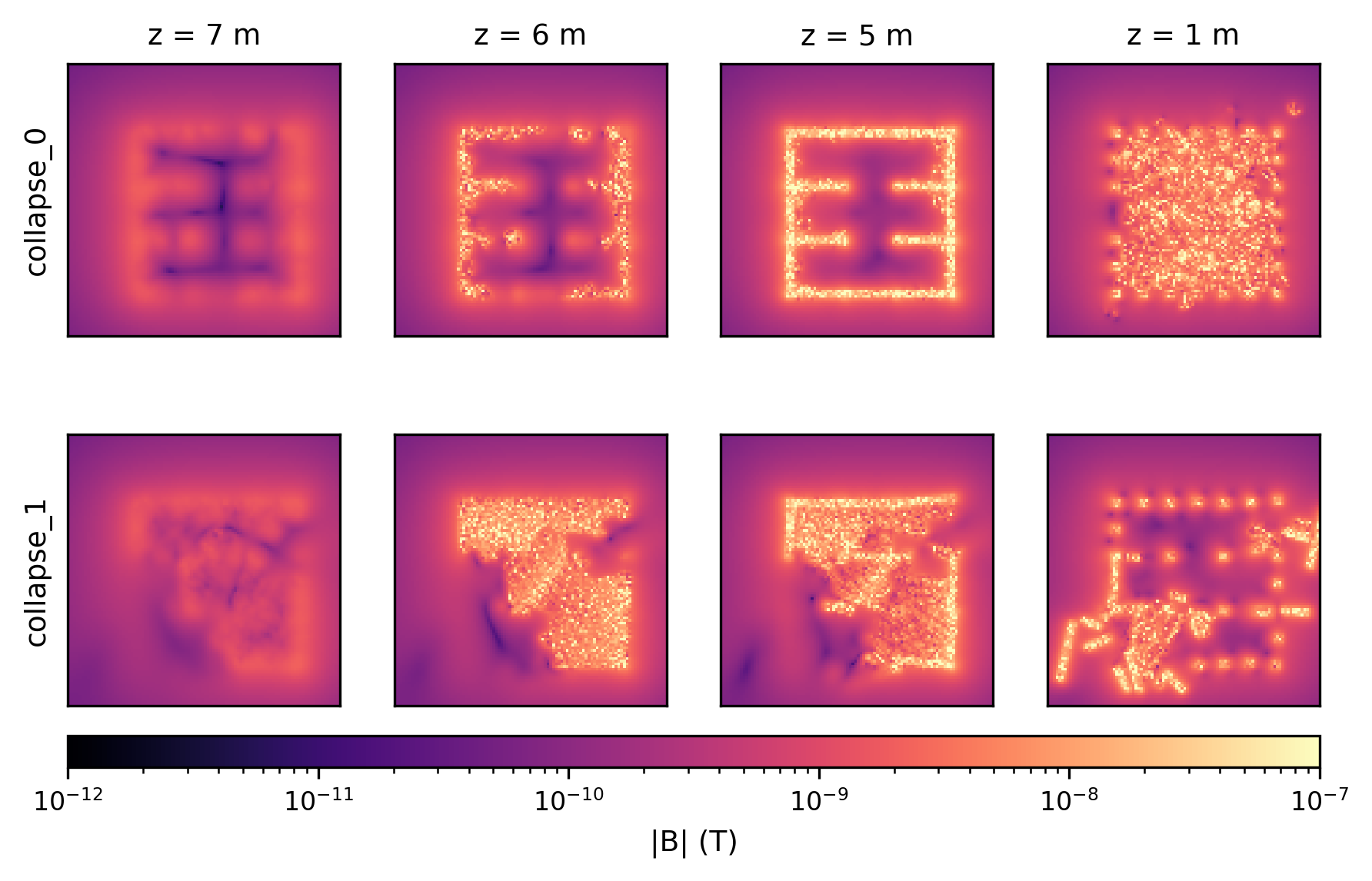}
    \caption{
        Horizontal field slices at $z = 7, 6, 5, 1$ m above ground for two independent collapse realizations (rows). 
        The standing structure is $5$ m tall in both cases.
    }
    \label{fig:z-cuts}
\end{figure*}

To set a quantitative sensor target we take horizontal slices of the field at four altitudes (Figure~\ref{fig:z-cuts}): 7 m and 6 m, the standoff regime $1$--$2$ m above the standing roof; 5 m, at the roof line; and 1 m, below the roof and inside the rubble. 
We repeat this for two independent collapse realizations to confirm the qualitative picture is not specific to a single mesh. 
At 6--7 m the building outline is clearly discernible in the sub-pT to sub-nT range for both meshes.
As the drone descends the outline sharpens, with strong clusters around the surviving pillars. 
Below the roof the field becomes noisy and the dipole approximation begins to break down. 
The takeaway is that the operating envelope for useful structural inference sits in the sub-pT to sub-nT range over the standoff distances of interest, well inside the published sensitivities of modern quantum magnetometer platforms~\citep{sengottuvel22a, herb25a, taylor08a}.

\subsection*{Autonomous mapping pipeline}
\begin{figure*}[t!]
    \centering
    \includegraphics[width=\linewidth]{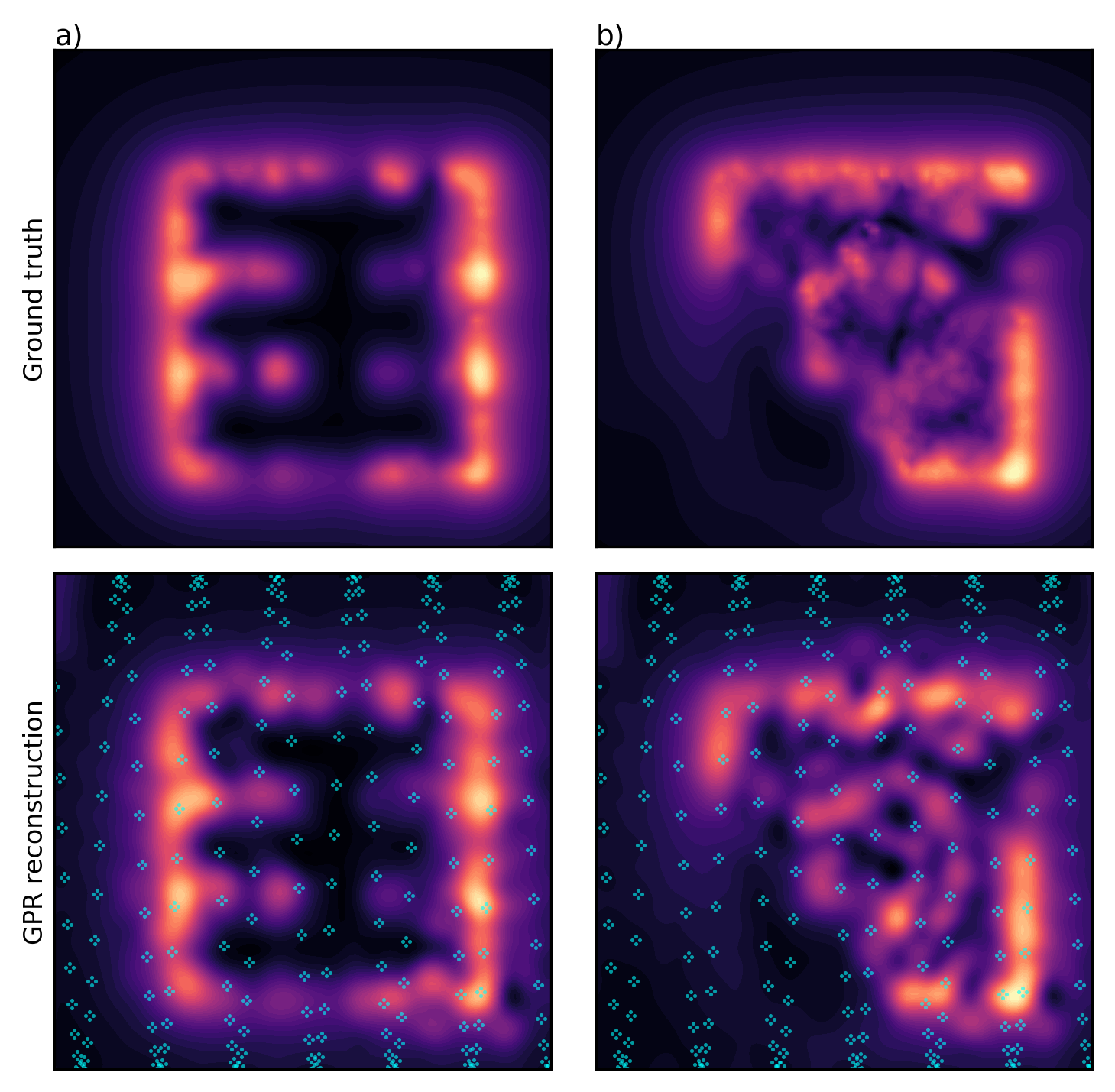}
    \caption{
        Sparse-sample GPR reconstruction over two collapse realizations, a) \texttt{collapse\_0} and b) \texttt{collapse\_1}. 
        Top row: dipole ground-truth field on the operating plane. 
        Bottom row: GPR reconstruction from a sine-wave flight, with the sampled waypoints overlaid in cyan. 
        The drone was modeled as flying at 7 m above ground, approximately 2 m above the building height.
    }
    \label{fig:mapping_demo}
\end{figure*}

Given a sensor that can resolve the field above, we now ask how efficiently a drone can recover the underlying structure. 
Figure~\ref{fig:mapping_demo} shows a single sine-wave flight over two distinct collapse realizations with a four-sensor array; the RBF-kernel GPR reconstruction (right column) recovers the gross structure of each rubble pattern from a few hundred sparse samples (cyan dots).

\begin{figure*}[h!]
    \centering
    \includegraphics[width=\linewidth]{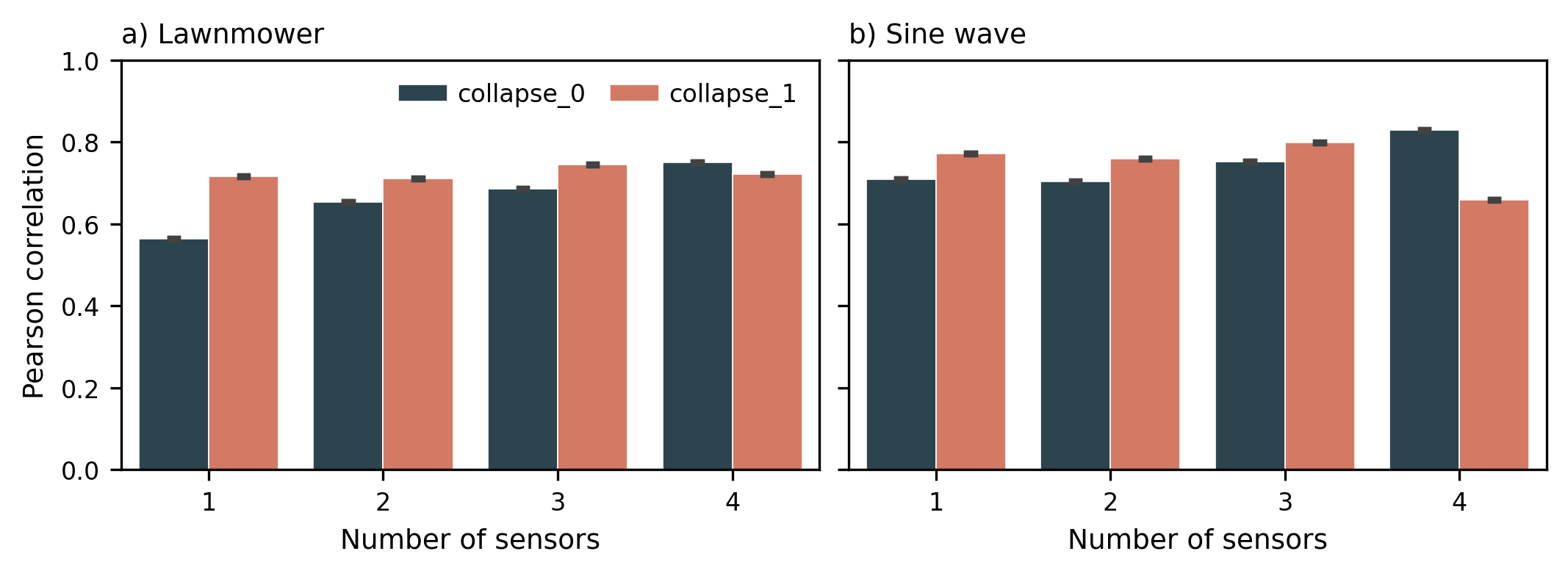}
    \caption{
        Pearson correlation between GPR reconstruction and ground truth as a function of sensor count for two independent collapse realizations, under a) lawnmower and b) sine-wave sweeps.
    }
    \label{fig:sensor_distribution}
\end{figure*}

The choice of array size $N$ trades off resolution against payload (Figure~\ref{fig:sensor_distribution}). For the structurally simpler \texttt{collapse\_0}, reconstruction quality improves monotonically through $N=4$ under both lawnmower and sine sweeps. 
For the more fragmented \texttt{collapse\_1} the improvement plateaus at $N=3$ and can regress at $N=4$ under the phase-sensitive sine trajectory. 
Taken together, $N=3$ emerges as the most robust choice across collapse geometries, with larger sensor arrays offering little additional advantage.

\begin{figure}[h!]
    \centering
    \includegraphics[width=\linewidth]{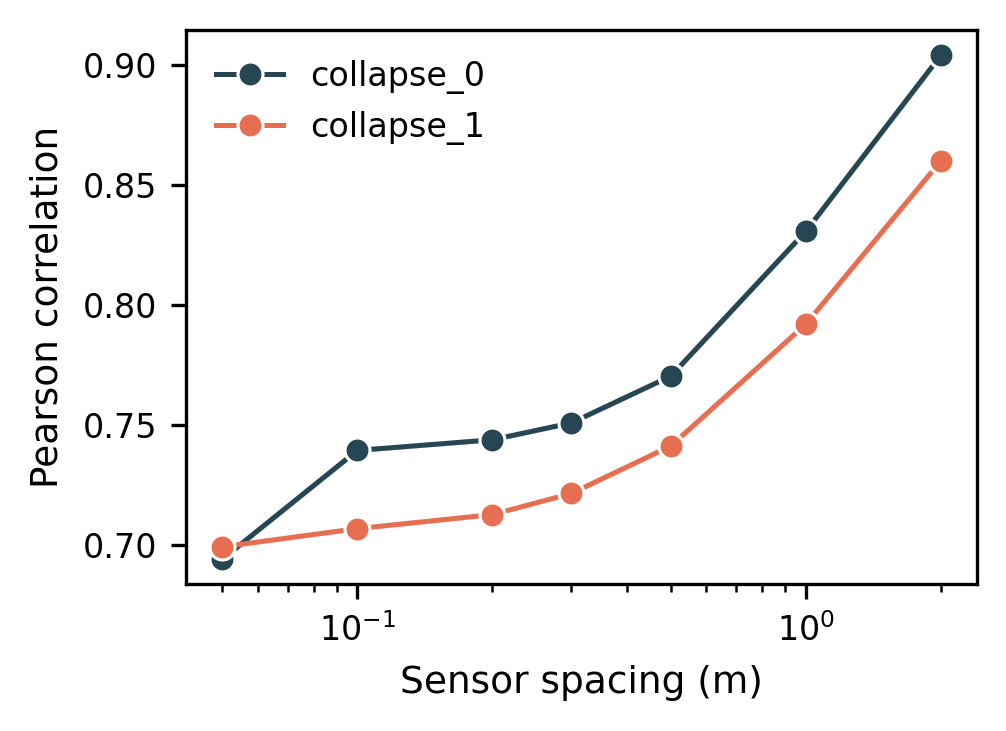}
    \caption{
        Pearson correlation between GPR reconstruction and ground truth as a function of sensor inter-element spacing for the $N=4$ diamond array, on log-spaced spacings from $5$ cm to $2$ m.
    }
    \label{fig:sensor_spacing}
\end{figure}

Beyond array size, the inter-element spacing of the sensors also shapes reconstruction quality (Figure~\ref{fig:sensor_spacing}). 
Pearson correlation grows monotonically with spacing for both meshes, from $\sim 0.70$ at $5$ cm to $\sim 0.90$ at $2$ m for \texttt{collapse\_0}. 
This is consistent with the role of a multi-sensor array as a local gradient estimator: a wider footprint per waypoint gives the GPR a stronger gradient signal, anchoring the reconstruction more firmly between flight lines. 
The practical trade-off is that a meter-scale sensor footprint is challenging to mount on a small UAV; the curve provides a quantitative basis for picking a spacing matched to a given platform.

\begin{figure*}[h!]
    \centering
    \includegraphics[width=0.9\textwidth]{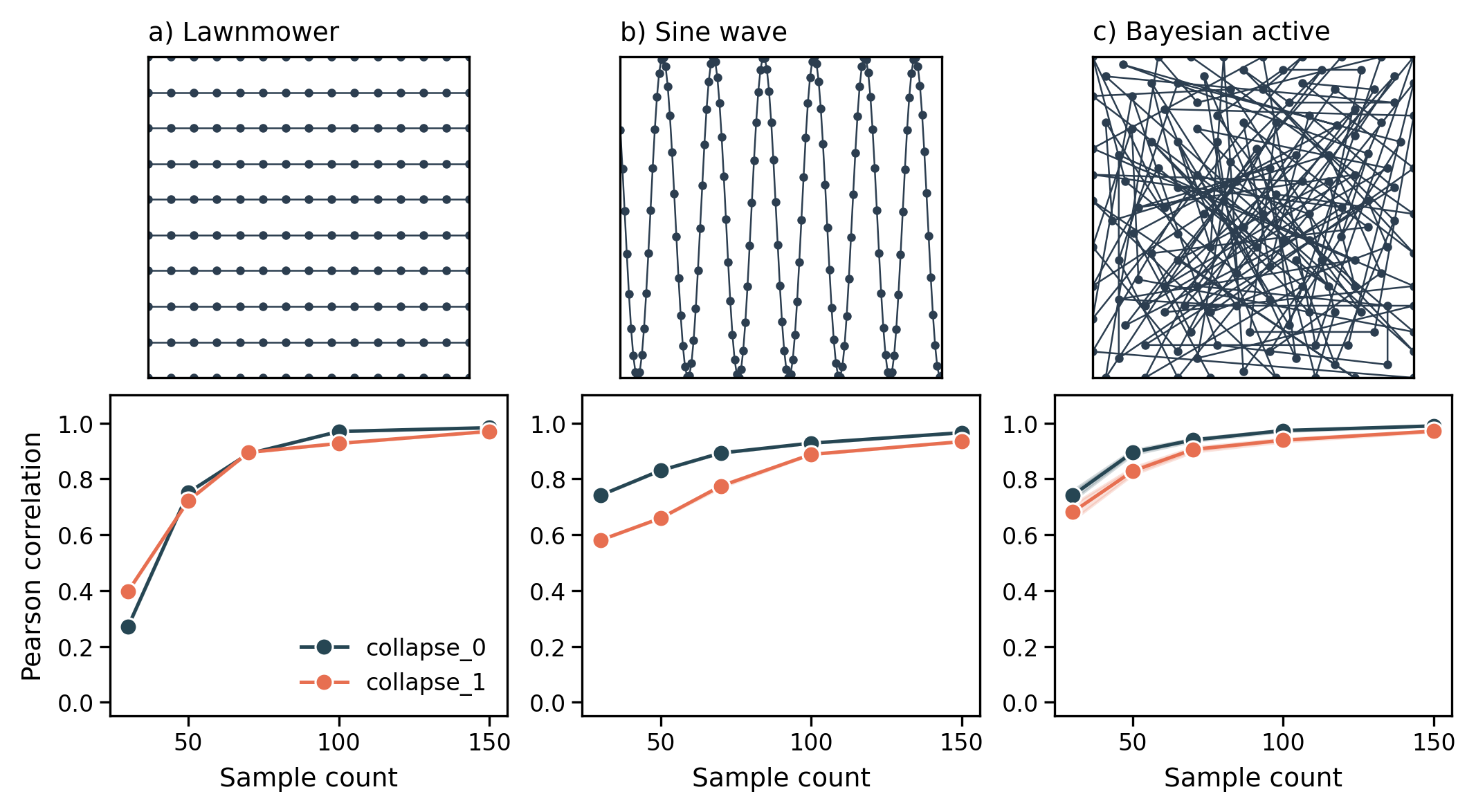}
    \caption{
        Sampling strategies as a function of sample budget. Top row: representative drone trajectory for a) lawnmower, b) sine wave, c) Bayesian active sampling. Bottom row: Pearson correlation between GPR reconstruction and ground truth at five sample budgets, with the Bayesian-active panel showing the genuine seed-to-seed variance (shaded band) from its $20$ random seed-point realizations.
    }
    \label{fig:sampling}
\end{figure*}

We compare deterministic and adaptive sampling strategies as a function of sample budget (Figure~\ref{fig:sampling}). 
The lawnmower converges slowly but reliably. 
The sine wave reaches high correlation faster but is sensitive to phase alignment with the underlying anomalies. 
The Bayesian-active sampler reaches peak correlation in roughly $100$ samples, well ahead of either deterministic strategy. 
Beyond saturation, the variance of the active sampler grows as the acquisition function increasingly directs the drone to low-signal regions at the boundary of the search area.

\section{Discussion}
\label{sec:discussion}
The results above sit on two foundations. 
The first is a physics claim: an induced-dipole model of steel-reinforced concrete rubble produces fields of order 30 pT to 3 $\mu$T at the source surface that fall by several orders of magnitude with standoff, leaving structurally informative contrast in the sub-pT to sub-nT range over the UAV operating regime and defining a quantitative sensitivity target.
The second is an algorithmic claim: a Gaussian-process reconstruction driven by Bayesian active sampling recovers the gross structure of the rubble from a handful of sparse drone samples.

\subsection*{Physical observability}
Figure~\ref{fig:full-map} shows that intact pillars and roof panels dominate the field while detached fragments contribute little, a pattern consistent with the per-triangle area scaling of the model and, physically, with rebar holding clusters of fragments together. 
Above the roofline the structure is visible from $\sim 1$ m of standoff with sub-pT to sub-nT contrast.
Below the roofline the noise from the small rubble pieces results in a noisy field.
Further, in this regime the dipole approximation begins to break down and a surface-integral formulation becomes a logical future step.

From an applications standpoint, the structural analysis implemented here could be used for void detection by mapping the continuous, magnetized wall structures and flagging gaps as candidate voids.
Closely related, the field map could also be used to plan safe rubble removal, which relies on understanding the residual structural integrity of damaged buildings~\citep{patel23a}, and high-sensitivity field maps of geological structures more broadly~\citep{abudeif25a}.

\subsection*{Algorithmic feasibility}
A three-sensor array is the size-weight-and-power optimal payload across the configurations tested: single-sensor reconstructions are unreliable while larger configurations offer diminishing gains. 
Among sampling strategies, deterministic lawnmower surveys give reliable but slow coverage, sine-wave sweeps are fast but phase-sensitive, and Bayesian active sampling reaches peak structural correlation in roughly $100$ samples, the regime relevant to time-critical search-and-rescue. Late-stage variance in the active sampler reflects the exploration-exploitation balance of the acquisition function: once the primary anomalies are resolved the agent is pushed outward into low-signal regions where additional samples add noise rather than information. 
A budget-aware acquisition that regularizes and terminates at map saturation is a natural extension.

\subsection*{Outlook}
The pipeline as it stands suggests two natural extensions.
First, the Bayesian acquisition selects only $(x, y)$ waypoints on a fixed operating plane; extending it to an altitude-aware acquisition over $(x, y, z)$ would let the drone fly higher and faster over homogeneous patches and drop down for finer-grained sampling over candidate structural anomalies, trading flight time against local resolution as the map develops.
Second, the reconstructed field map is itself an actionable input to the downstream tasks already discussed: void detection, i.e., identifying empty, non-magnetic structures within the rubble for survivor search-and-rescue, and structural-integrity assessment for safe rubble clearance.
Both will benefit from an automated void-detection layer that operates directly on the GPR-reconstructed field, converting a continuous map into a discrete set of high-probability void locations to be handed off to a ground-based robotic search component.

\section{Conclusion}
\label{sec:conclusion}
We have presented a simulated pipeline for drone-based quantum-magnetometry of disaster scenes. 
A per-triangle dipole model on Unreal Engine rubble shows that meaningful structure is recoverable from steel-reinforced concrete in the sub-pT to sub-nT range at standoff distances of one to two meters.
This range sits firmly inside the sensitivity envelope of modern NV-center magnetometers.
A Gaussian-process reconstruction with a three-sensor array and Bayesian active sampling recovers the gross structure of the rubble in roughly $100$ samples, with the conclusion holding across multiple independent collapse realizations.
Future work may replace the dipole approximation with a gradiometric formulation for ground-based operations, fold measurement noise and finite $T_2^{*}$ effects into a calibrated end-to-end sensor model, and add battery-aware termination to the active sampler. 
The results indicate that both halves of the problem, sensing and reconstruction, are within reach of current platforms.

\section*{Code and Data Availability}
\label{sec:code-availability}
The simulation pipeline, Unreal Engine simulations, and collapse meshes described in this work can be made available upon reasonable request.

\section*{Acknowledgments}
The authors acknowledge helpful conversations with Hideaki Yoshimura, Yaswitha Gujju, and Sophie Colleen Stearn. 
This work was completed as a contribution to the NEDO Challenge Quantum Computing “Solve Social Issues!”.

\bibliography{bibliography}

\end{document}